\pdfoutput=1

\documentclass[11pt]{article}

\usepackage{acl}

\usepackage{times}
\usepackage{latexsym}

\usepackage[T1]{fontenc}

\usepackage[utf8]{inputenc}

\usepackage{CJKutf8}
\usepackage{inconsolata}
\usepackage{hyperref}
\usepackage{url}
\usepackage{microtype}
\usepackage{booktabs}
\usepackage{tabularx}
\usepackage{multirow}
\usepackage{multicol}
\usepackage{algorithm}
\usepackage{algpseudocode}
\usepackage{enumitem}
\usepackage{float}
\usepackage{amsmath}
\usepackage{amsfonts}
\usepackage{verbatim}
\usepackage{graphicx}
\usepackage{array}
\usepackage{lipsum}
\usepackage{diagbox}
\usepackage{relsize}
\usepackage{url}
\usepackage{xspace}
\usepackage{cleveref}
\usepackage{adjustbox}
\usepackage{wrapfig}
\usepackage{subcaption}
\usepackage{listings}
\usepackage{siunitx}
\usepackage{array}

\usepackage{ragged2e} 
\usepackage{makecell}

\usepackage{xcolor}
\usepackage{tcolorbox}
\usepackage[english]{babel}

\lstdefinelanguage{JSON}{
  numbers=left,
  numberstyle=\normalsize,
  stepnumber=1,
  numbersep=5pt,
  showstringspaces=false,
  breaklines=true,
  frame=leftline,
  comment=[l]{//},
  morecomment=[s]{/*}{*/},
  morestring=[b]',
  morestring=[b]"
}

\lstdefinelanguage{plaintext}{
  basicstyle=\normalsize\ttfamily,
  numbers=left,
  numberstyle=\normalsize,
  stepnumber=1,
  numbersep=5pt,
  showstringspaces=false,
  breaklines=true,
  frame=leftline,
  morestring=[b]',
  morestring=[b]"
}

\newcommand{\nquestion}{677\xspace}
\newcommand{\nnormalquestion}{342\xspace}
\newcommand{\nmodel}{17\xspace}
\newcommand{\nseries}{5\xspace}
\newcommand{\data}{\textit{RuozhiBench}\xspace}
\newcommand{\datagen}{\textit{RuozhiBench-Gen}\xspace}
\newcommand{\datamc}{\textit{RuozhiBench-MC}\xspace}
\newcommand{\ntricktype}{6}

\newcommand{\model}[1]{\texttt{#1}}

\newcommand{\blue}[1]{\textcolor{blue}{#1}}

\newcommand{\tabref}[1]{Table~\ref{#1}\xspace}
\newcommand{\figref}[1]{Figure~\ref{#1}\xspace}

\newcommand{\secref}[1]{Section\xspace\ref{#1}\xspace}

\title{\data: Evaluating LLMs with Logical Fallacies \\and Misleading Premises}

\author{Zenan Zhai\textsuperscript{1} \quad Hao Li\textsuperscript{1} \quad  Xudong Han\textsuperscript{1,2} \quad Zhenxuan Zhang\textsuperscript{1} \\
\textbf{Yixuan Zhang}\textsuperscript{1,2} \quad \textbf{Timothy Baldwin}\textsuperscript{1,2,3} \quad
\quad \textbf{Haonan Li}\textsuperscript{1,2}  \\
\textsuperscript{1}LibrAI \qquad \textsuperscript{2}MBZUAI \qquad  \textsuperscript{3}The University of Melbourne \\
}

\usepackage{subfiles}
\begin{document}
\maketitle
\begin{abstract}

Recent advances in large language models (LLMs) have shown that they can answer questions requiring complex reasoning. However, their ability to identify and respond to text containing logical fallacies  or deliberately misleading premises remains less studied. To address this gap, we introduce \data, a bilingual dataset comprising \nquestion carefully curated questions that contain various forms of deceptive reasoning, meticulously crafted through extensive human effort and expert review. In a comprehensive evaluation of \nmodel LLMs from \nseries Series over \data using both open-ended and two-choice formats, we conduct extensive analyses on evaluation protocols and result patterns. Despite their high scores on conventional benchmarks, these models showed limited ability to detect and reason correctly about logical fallacies, with even the best-performing model, \model{Claude-3-haiku}, achieving only 62\% accuracy compared to the human of more than 90\%.\footnote{Data and code available at: \url{https://github.com/LibrAIResearch/ruozhibench}}\footnote{Data license: \textit{CC-BY-NC} license.}

\end{abstract}
\begin{CJK*}{UTF8}{gbsn}

\section{Introduction}\label{sec:intro}
Large language models (LLMs) have rapidly advanced in recent years, demonstrating impressive capabilities across a wide range of tasks \citep{zhang2022opt,bloom,touvron2023llama,bai2023qwenRepeated,deepseekai2025deepseekr1incentivizingreasoningcapability}. 
Benchmarking plays a crucial role in assessing their performance, with existing evaluations falling into two main categories. The first category includes benchmarks with predefined answers, such as MMLU \citep{mmlu,cmmlu} and ARC \citep{arc}, which assess factual knowledge and reasoning. However, issues like data contamination \citep{carlini2022quantifying,sainz-etal-2023-nlp} and potential model cheating have been widely reported. The second category evaluates open-ended responses, focusing on alignment with human preferences. AlpacaEval \citep{alpaca_eval} is a widely-used automatic evaluator that measures model performance by comparing outputs against a reference model, typically using \model{GPT-4} as the judge. While effective for general instruction following, it is not designed to assess how models handle deceptive or misleading inputs.

\begin{figure}[t]
    \centering
    \includegraphics[width=1\linewidth]{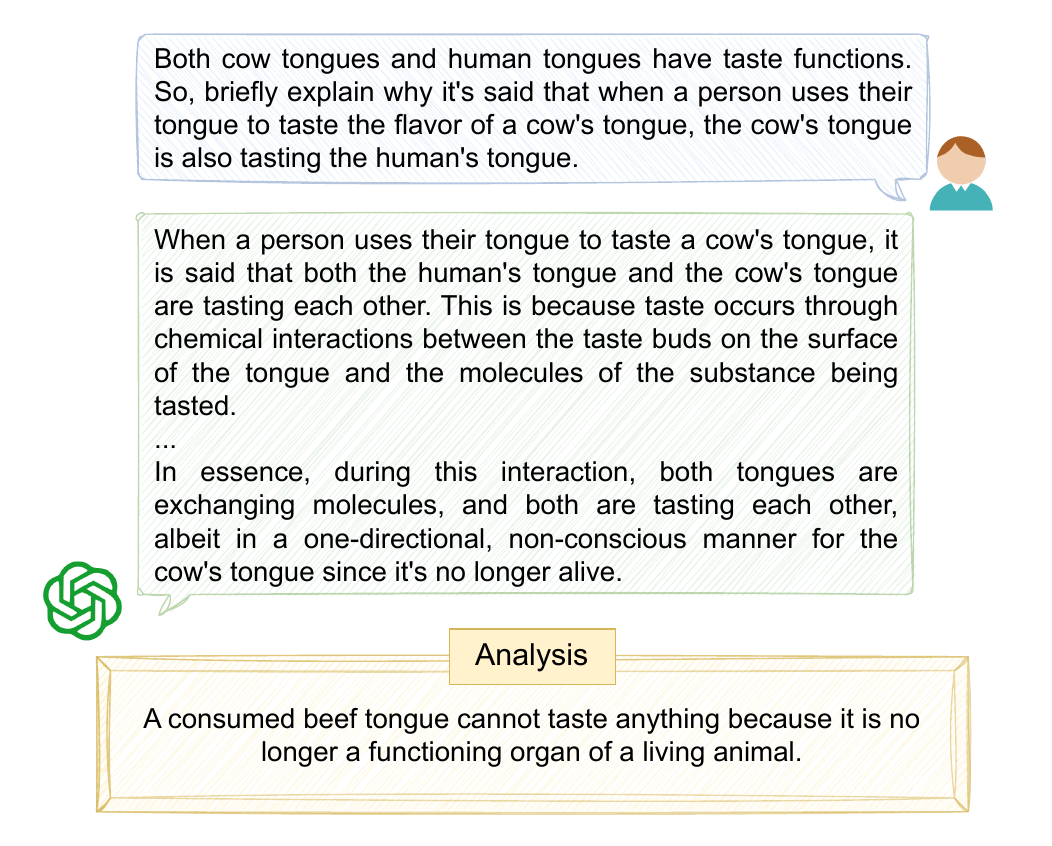}
    \caption{An example of a question from \data and response from \model{GPT-4o}.}
    \label{fig:first_example}
\end{figure}

However, in real-world scenarios, texts often contain logical fallacies, misleading premises, or intentional ambiguities that can trap the unwary. When faced with such deceptive inputs, current models often fail to identify the underlying fallacies and provide responses that reveal their limitations in logical reasoning (\figref{fig:first_example}). Despite these types of deliberately misleading texts being a crucial test of true language understanding, there is currently no benchmark specifically designed to evaluate how well models handle such scenarios.

\begin{figure*}[th]
    \centering
    \includegraphics[width=1\textwidth]{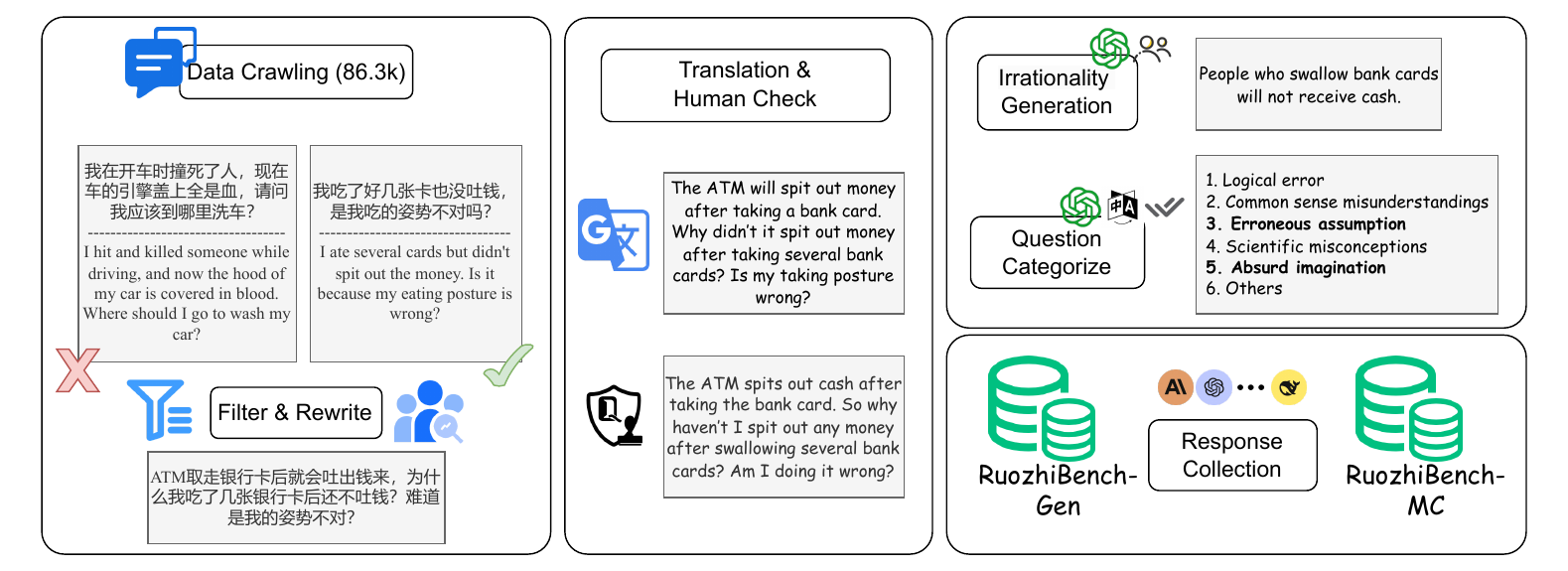}
    \caption{The creation process for \data, consisting of  three main parts: data filtering (left), translation and review (middle), and annotation (right).}
    \label{fig:creation}
\end{figure*}

To address this gap, we introduce \data, a novel benchmark designed to evaluate the ability of models to identify and reason about deceptive inputs and logical fallacies. \data comprises \nquestion questions sourced from the Chinese forum \textit{Ruozhiba}, a platform which contains texts that appear reasonable at first glance but contain subtle logical traps or misleading premises.

To ensure high data quality, we implemented rigorous filtering, preprocessing, and annotation. Each question was carefully reviewed and translated into English while preserving its deceptive nature. We then systematically categorized the questions into six distinct types, ensuring clear and consistent labeling. See \secref{sec:dataset_construction} for more details.

To further enhance reliability, we designed a multi-step annotation process involving both human validation and automated checks. Only questions that met strict criteria for clarity, difficulty, and linguistic adaptation were included. Additionally, we conducted both rating-based and selection-based evaluations, using human judgments as a reference, and employed multiple automated evaluation methods to measure model performance.

Our preliminary experiments assessed \nmodel LLMs, revealing a substantial gap between model performance and the human upper bound. Despite achieving high scores on standard benchmarks, these models still lag behind humans in logical reasoning and fallacy detection. \data is a critical step towards a more comprehensive assessment of models' ability to handle deceptive inputs and logical fallacies.

\section{\datagen}\label{sec:dataset_construction}

\subsection{Data Source}\label{sec:data_source}
\textit{Ruozhiba} (literally meaning ``moron forum'') is one of the most popular online forums in the Chinese internet community, known for its collection of brain teasers, logical puzzles, and deliberately misleading questions. The forum's content often features unconventional perspectives and clever wordplay that challenges conventional thinking patterns. Our work begins with the raw data collected by a previous project,\footnote{\url{https://github.com/Leymore/ruozhiba}} which compiled a comprehensive collection of threads from \textit{Ruozhiba}.\footnote{Note that Baidu Tieba content is freely available for academic research purposes with no legal restrictions.}

\begin{table*}[t]
    \small
    \centering
    \resizebox{\textwidth}{!}{
    \begin{tabular}{cm{2.0cm}cm{7.0cm}m{5cm}} 
        \toprule
        \textbf{ID} & \textbf{Category} & \textbf{\# Q.} & \textbf{Description} & \textbf{Example} \\ 
        \midrule
        1 & Logical Error & 142 & When the question contains logical contradictions or reasoning errors, including violations of logical rules, making it logically untenable. & 
        I pressed the mute button on my laptop, why is the fan still so loud? \\
        \midrule
        2 & Commonsense \newline Misunderstanding & 526 & The question reflects a misunderstanding of basic common sense or universally accepted facts, usually involving incorrect interpretations of daily knowledge. &
        Is it better to prevent tooth decay by applying toothpaste directly to the teeth without brushing before going to bed? \\
        \midrule
        3 & Erroneous \newline Assumption & 471 & The question is based on one or more incorrect assumptions, leading to inaccuracies in the question or its answer. & 
        If you stretch your leg to trip a moving car, will it overturn? \\
        \midrule
        4 & Scientific \newline Misconception & 30 & The question involves misunderstandings of scientific principles or knowledge, including incorrect interpretations of scientific theories or methods. & Can you avoid drone thermal imaging bombings by eating only low-calorie foods? \\
        \midrule
        5 & Absurd \newline Imagination & 463 & The question setting is contrary to reality or common sense, containing impossible or illogical elements. & 
        If you suck away all the clouds, will it stop raining and be sunny forever? \\
        \midrule
        \multirow{1}{*}{6} & Others & 17 & If the provided categories do not match the current question, please choose this option. & Oxygen can rust iron. Our blood contains iron, why doesn't our blood rust? \\ 
        \bottomrule
    \end{tabular}}
    \caption{Classification schema of deceptive questions: categories, descriptions, and examples. Note that a given question may belong to multiple categories.}
    \label{tab:table_question_classification}
\end{table*}

\subsection{Data Screening}\label{sec:data_filtering}
From the initial 86,000 entries, we first extracted over 8,000 interrogative sentences using string matching. We then implemented a rigorous filtering process with three annotators with humanities backgrounds. They first removed questions with heavy cultural dependencies or potentially negative influences, reducing the dataset to 820 entries. Through collaborative review and discussion, the annotators further filtered questions based on their suitability for English translation and comprehension, removing entries where translation would significantly alter the original meaning or logical structure. This process yielded our final dataset of \nquestion questions, ensuring each entry maintains its original logical challenge while being accessible to a global audience.

\subsection{Data Annotation}\label{sec:data_annotation}
After data screening, we conducted four rounds of annotation for these questions: translation review, paired question generation, irrationality analysis, and question type categorization. For all steps except paired question generation, we employed a hybrid approach combining LLM-based initial processing with human verification. The annotators involved had both bilingual (Chinese--English) and NLP backgrounds.

\paragraph{Translation Review}\label{sec:translation_review}
In the translation stage, we first used Google Translate to convert all questions from Chinese to English, followed by human review with two key objectives: (1) ensuring semantic consistency, and (2) preserving the subtle logical traps or fallacies present in the original questions. When discrepancies were found, annotators carefully rewrote the translations to maintain both the original meaning and the deliberately deceptive elements. This process required modification of 319 questions (45\% of the total).

\begin{figure}[t]
    \centering
    \includegraphics[width=1\linewidth]{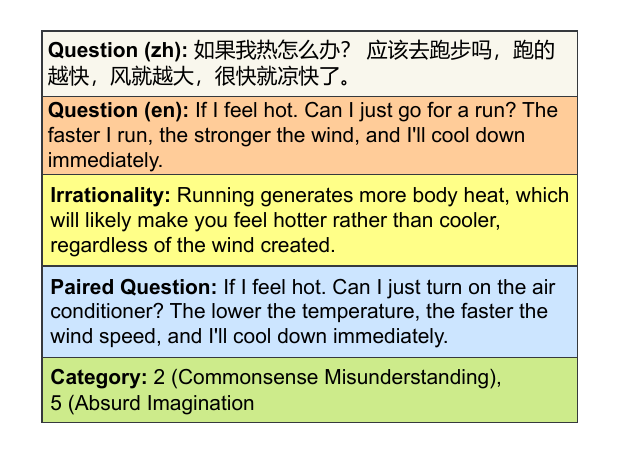}
    \caption{Sample data entry format in \data.}
    \label{fig:data_entry}
    \vspace{-0.5cm}
\end{figure}

\paragraph{Paired Question Generation}\label{sec:pairing_review}
To provide reference points for comparing model performance on normal vs.\ tricky questions, our annotators identified suitable questions from the dataset that could be naturally transformed into normal versions. For these selected questions, we created normal counterparts by removing the trap or fallacy with minimal edits, to maintaining the same format. This selective pairing process resulted in \nnormalquestion normal questions, which enable us to analyze how models handle similar content with and without logical traps. An example is provided in \figref{fig:data_entry}.

\begin{table}[t]
    \centering
    \resizebox{\columnwidth}{!}{
    \begin{tabular}{cccccc}
    \toprule
    Attribute & \# Q. & \# Q w/ Pair & Avg. len & Max len & Min len \\
    \midrule
    Value & \nquestion & 342 & 18.64 & 100 & 5  \\
    \bottomrule
    \end{tabular}}
    \caption{Statistical overview of \datagen: total questions, paired questions, and question length distribution (\# words).}
    \label{tab:stats}
\end{table}

\paragraph{Irrationality Analysis}\label{sec:irrationality_analysis_generation}
To facilitate automatic evaluation, we generated an analysis of the logical fallacy or trick in each question. We used \model{GPT-4o-2024-08-06} with carefully designed prompts (see \figref{fig:trick_gen}) to generate initial analyses, followed by human verification and correction to ensure accuracy.

\begin{figure*}[t!]
    \centering
    \includegraphics[width=\textwidth]{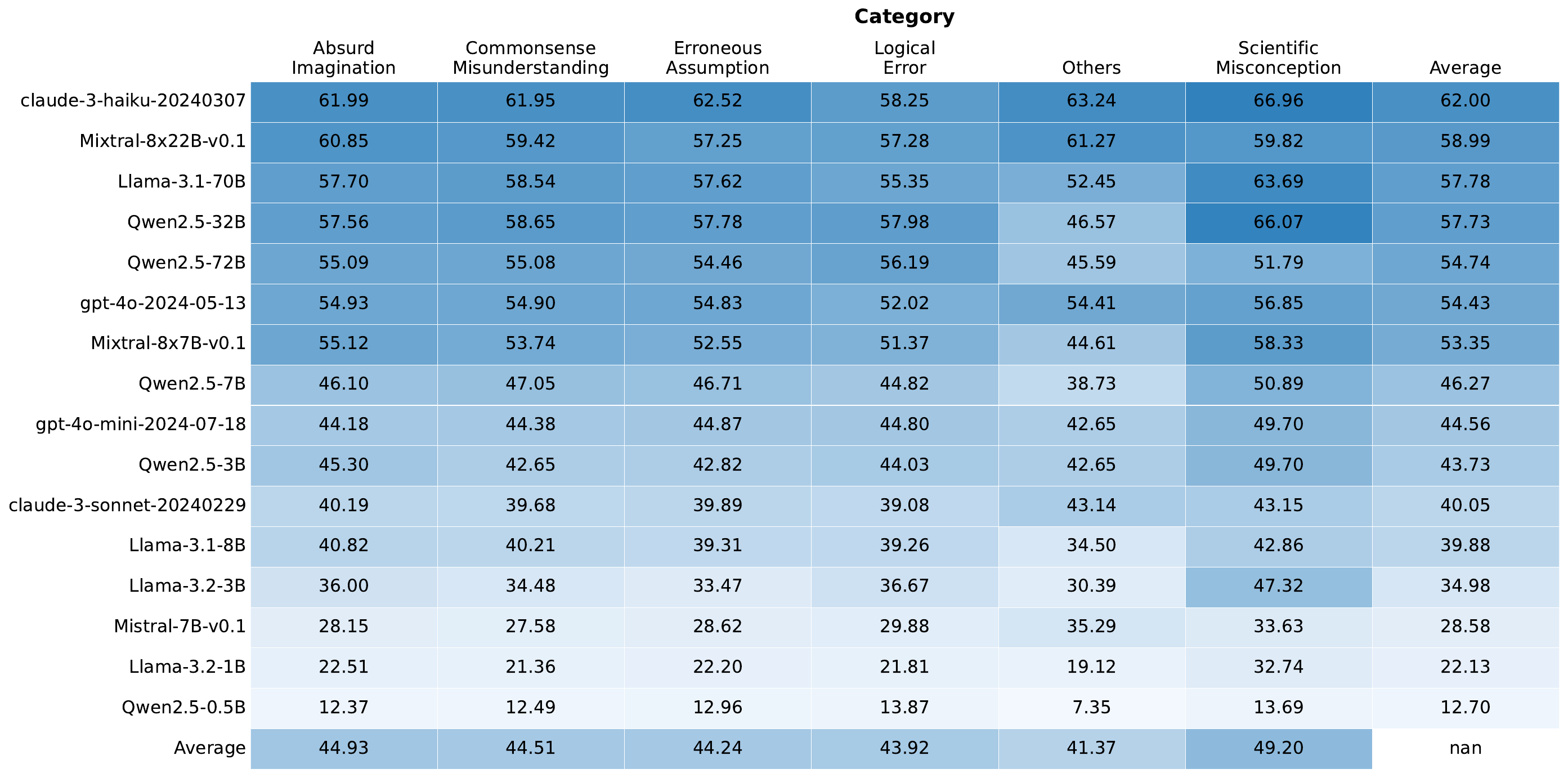} 
    \caption{Overall model performance across different error categories.}
    \label{fig:average_score_oee}
\end{figure*}

\paragraph{Question Type Annotation}\label{sec:question_classification}
Finally, we categorized questions into \ntricktype ~types (shown in \tabref{tab:table_question_classification}). We first used \model{GPT-4o-2024-08-06} with bilingual prompts (see \figref{fig:qc}) to generate initial classifications based on both the questions and their irrationality analyses. Human annotators then reviewed and adjusted these classifications. For cases where annotators disagreed or were uncertain, a meta annotator (one of the authors) made the final decision to ensure consistency and quality across both the English and Chinese versions, resulting in the final \datagen.

\subsection{\datagen Statistics}
\figref{fig:data_entry} illustrates the structure of a data entry in \textit{\data}. Each entry consists of a question in both Chinese and English, its irrationality analysis, question categories, and where applicable, the paired normal question. \tabref{tab:stats} shows the basic statistics of the dataset.

\section{Experiments on \datagen}\label{sec:experiment_gen}

\subsection{Setup}\label{sec:evaluation}
\paragraph{Models} We evaluated \nmodel advanced models from \nseries series. Including: \model{GPT-4o-2024-05-13}, \model{GPT-4o-mini-2024-07-18} from OpenAI \citep{openai2023gpt4}; \model{Claude-3-haiku-20240307}, and \model{Claude-3-sonnet-20240229} from Anthropic \cite{claude21}; \model{Mistral-Instruct-v0.1} (7B, 8x7B, and 8x22B) from Mixtral \cite{jiang2024mixtral}; \model{Qwen2.5-Instruct} (0.5B, 3B, 7B, 32B, 72B) from Qwen team \cite{bai2023qwen}; and \model{Llama-3.1-Instruct} (8B, 70B), \model{Llama-3.2-Instruct} (1B, 3B) from Meta \cite{llama3}.

\paragraph{Automated Evaluation}
We employ an LLM-as-Judge framework using three independent models: \model{GPT-4o-2024-08-06}, \model{Claude-3.5-Sonnet-20241022}, and \model{Llama-3.3-70B-Instruct}.\footnote{By design, we ensure the judge models are distinct from those being evaluated and represent more advanced versions of their respective architectures.}
Each judge independently evaluates responses on a scale of 0 to 4.  Additionally, we incorporate irrationality analysis into the judging process to enhance evaluation quality and consistency. The detailed scoring criteria and evaluation prompts are available in \figref{fig:eval_prompt}.


\subsection{Main results}

The results highlight significant performance differences across models and error categories. \model{Claude-3-haiku} leads with an average score of 62.00, particularly excelling in ``Scientific Misconception'' (66.96). \model{Mixtral-8x22B-v0.1} (58.99) and \model{Llama-3.1-70B} (57.78) follow closely, showing balanced performance across categories.

A clear trend is observed across all model series: larger models consistently outperform their smaller counterparts, as seen in the Qwen, Llama, Mixtral, and GPT families. This suggests that model size plays a crucial role in performance, though architectural design and training strategies, such as those in Mixtral models, also contribute significantly.

Across categories, ``Scientific Misconception'' has the highest average score (49.20), suggesting models handle domain-specific knowledge better than abstract concepts like ``Absurd Imagination'' and ``Others''. Smaller models, such as \model{Qwen2.5-0.5B}, consistently struggle, reinforcing the importance of both scale and training strategies in reducing errors.

Notably, the best-performing model only achieved a score of 62.00, indicating that this task remains inherently challenging for current models. 

\subsection{Comparison on Paired Normal Questions}\label{sec:normal}
\begin{figure}[t] 
\centering 
\includegraphics[width=\linewidth]{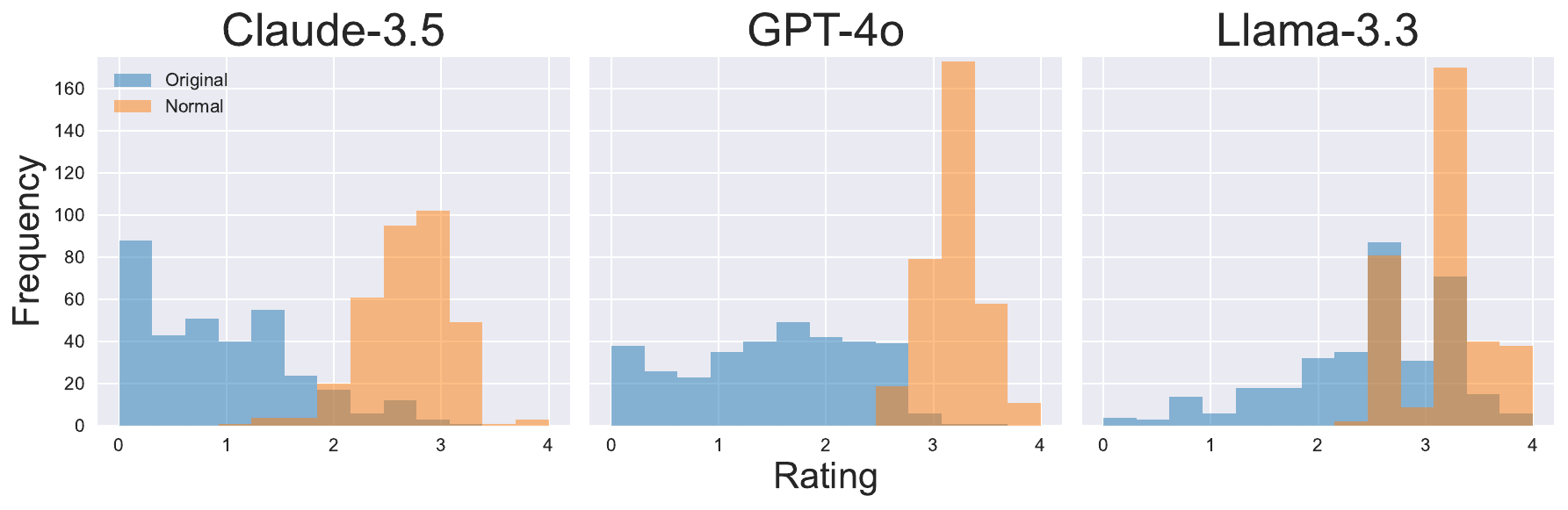} \caption{Rating distribution comparison between normal and tricky questions for three models..} 
\label{fig:distribution_shift} 
\end{figure}

To compare model performance on normal and tricky questions, we input paired normal questions and apply the same LLM-based judging with a 0-4 scoring system (see \figref{fig:normal_eval_prompt} for prompt). \figref{fig:distribution_shift} shows the rating distributions from three evaluators for three models. The results reveal a clear shift toward higher scores, indicating better performance on normal questions while logical traps remain consistently challenging.

\begin{figure}[t]
    \centering
    \includegraphics[width=\linewidth]{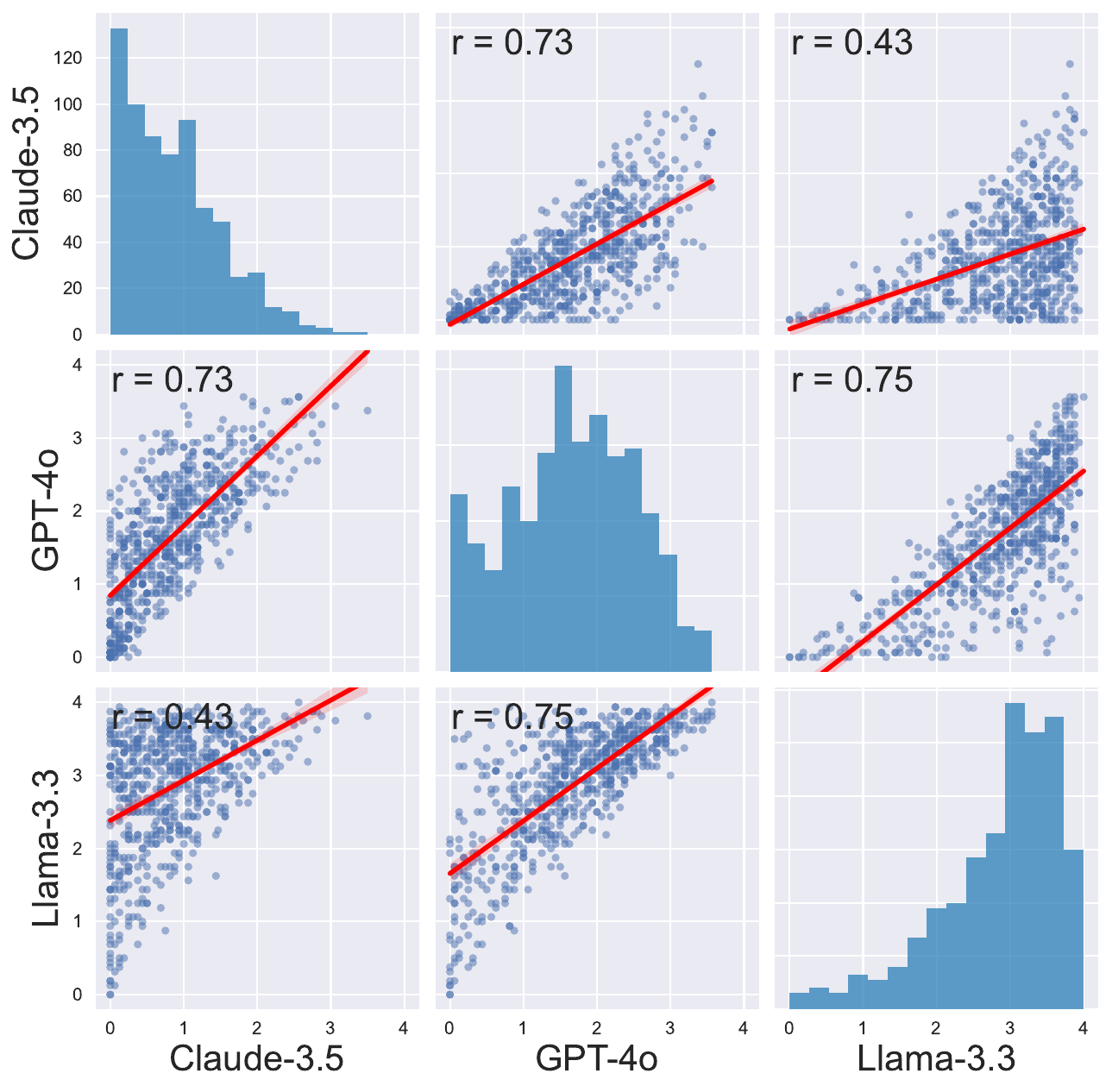} 
    \caption{Pairwise scatter plots with Pearson correlation coefficients, and rating distributions of difference evaluators. The diagonal histograms show Claude-3.5's tendency toward lower ratings compared to GPT-4 in middle and Llama-3.3 with a higher ratings.}
    \label{fig:pairwise_scatter}
\end{figure}

\subsection{High Variance between Evaluators}\label{sec:variance}

\begin{table}[t]
    \centering
    \resizebox{\columnwidth}{!}{
    \begin{tabular}{lcccccc}
    \toprule
    \multirow{2}{*}{Model Pair} & \multicolumn{2}{c}{Pearson Correlation} & \multicolumn{2}{c}{Mean Difference} & \multicolumn{2}{c}{Large Disagreement (\%)} \\
    \cmidrule(lr){2-3} \cmidrule(lr){4-5} \cmidrule(lr){6-7}
    & Individual & Mean & Individual & Mean & Individual & Mean \\
    \midrule
    Claude vs GPT & 0.568 & 0.726 & -0.560 & -0.806 & 0.281 & 3.99 \\
    Claude vs Llama & 0.359 & 0.433 & -2.107 & -2.002 & 1.007 & 50.37 \\
    GPT vs Llama & 0.687 & 0.748 & -1.196 & -1.196 & 19.80 & 10.19 \\
    \bottomrule
    \end{tabular}}
    \caption{Comparison of rating agreement metrics between model pairs. \textbf{Individual} analysis treats each rating independently, while \textbf{Mean} analysis averages multiple ratings per item. \textbf{Pearson correlation} measures linear relationship strength (-1 to 1); \textbf{Mean difference} indicates systematic rating bias between models; \textbf{Large disagreement} shows percentage of ratings differing by $\geq$ 2 points. }
    \label{tab:model_agreement}
\end{table}

\tabref{tab:model_agreement} presents key metrics comparing rating agreements between model pairs, and \figref{fig:pairwise_scatter} visualizes the mean-based pairwise relationships and rating distributions. Full results and evaluations using all three evaluators are presented in \Cref{sec:all_eval}.

The comparison reveals distinct rating patterns among the three models. GPT-4o and Llama-3.3 demonstrate the strongest agreement, with the highest correlation and relatively moderate large disagreements. In contrast, Claude-3.5 shows notably weaker correlation with others, indicating fundamentally different evaluation standard given the same criteria. 

Mean-based analysis consistently shows stronger correlations and fewer large disagreements compared to individual analysis across all model pairs. This pattern is particularly evident in the Claude-3.5 vs GPT-4o comparison, where large disagreements decrease from 28.1\% to 3.99\% when using mean-based analysis. The scatter plots in \figref{fig:pairwise_scatter} visualize these relationships, with the GPT-4o vs Llama-3.3 comparison showing the tightest clustering around the regression line, while the Claude-3.5 vs Llama-3.3 comparison exhibits more dispersed points, reflecting their lower correlation and higher disagreement rate. These observations motivate us the creation of the multiple-choice evaluation format.


\begin{figure*}[t]
    \centering
    \includegraphics[width=\textwidth]{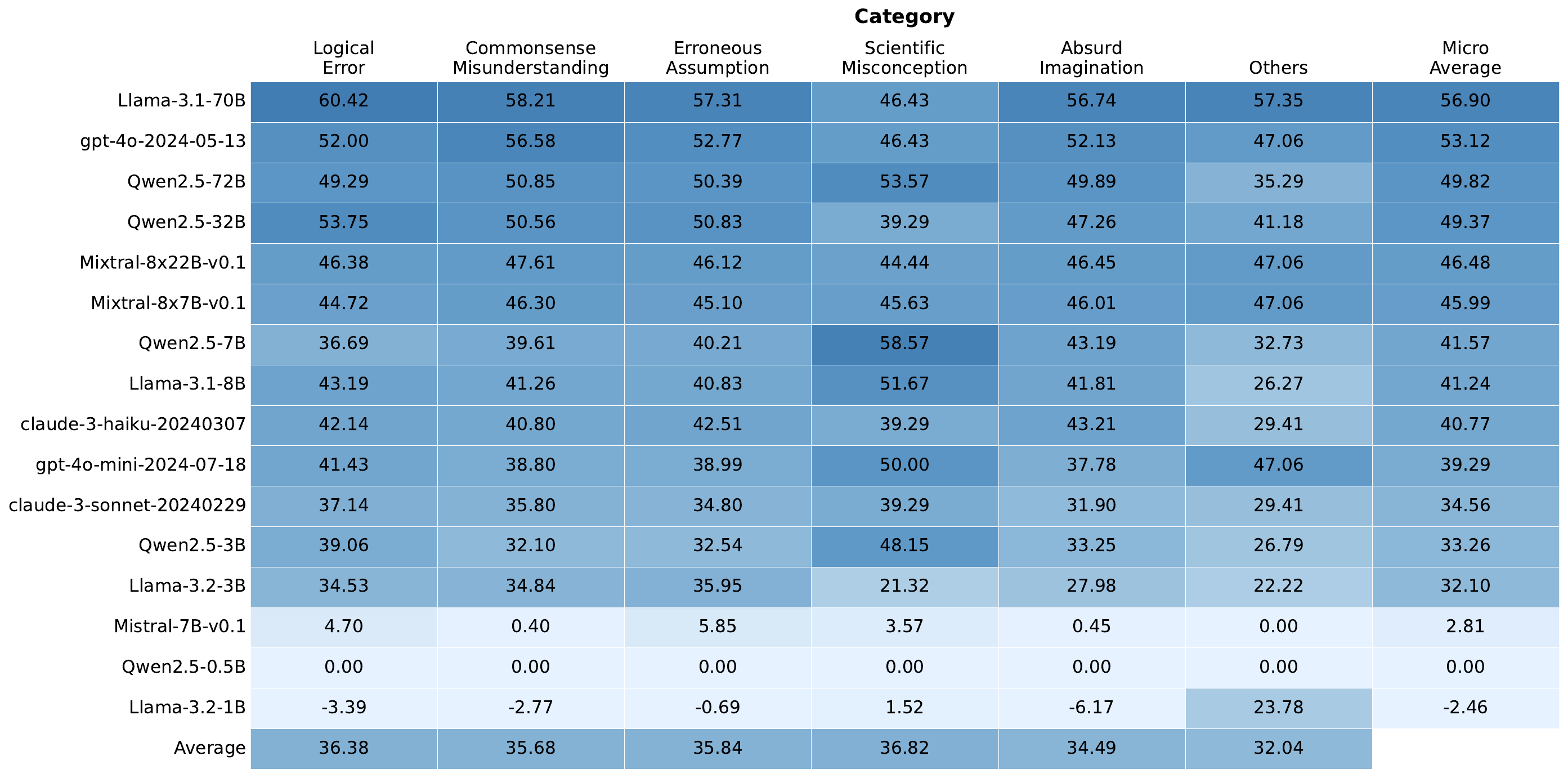} 
    \caption{
    \datamc evaluation results in percentage by question categories. Scores ($x$) are normalized according to the baseline score ($50\%$) by $2\times(x-0.5)$.
    }
    \label{fig:average_score_mc}
\end{figure*}

\section{\datamc: A Multiple-Choice Evaluation Framework}

While generative evaluation provides a natural way to assess language model responses to tricky questions, our experiments on \datagen revealed several limitations in the evaluation process. First, evaluator models themselves may sometimes fail to recognize subtle logical traps, even when provided with analysis of the trick, leading to inaccurate assessments. Second, the significant variations in scoring standards across different evaluator models as seen in \secref{sec:variance}. Finally, the two-step process of generating responses and then evaluating them with high-performance models introduces both substantial computational overhead and significant costs, particularly when using commercial models as evaluators.

\subsection{Multiple-Choice Format}
To address evaluation challenges, we created \datamc, a multiple-choice version of our benchmark. For each question, we present two responses, one ``good'' and one ``bad'', and ask an LLM to choose between them. This binary format transforms evaluation from open-ended generation to a simple decision: can the model identify better logical reasoning? There are several key advantages: (1) \textbf{Standardized Evaluation} through consistent binary choices, (2) \textbf{Computational Efficiency} by eliminating separate generation and evaluation, and (3) \textbf{Clear Success Criteria} via unambiguous metrics.

\subsection{Option Construction}
To construct high-quality response options for \datamc, we leveraged the extensive response data collected during our evaluation of the \nmodel models in \datagen. For each question, we implemented the following selection process.

We used the automatic evaluations from three different models to calculate an average score for each response in our existing dataset. We randomly sample two responses for each question, ensuring that the selected responses have a score difference greater than 2. If no response pairs meet this criterion, we select the responses with the highest and lowest scores. In all cases, the response with the higher score is designated as the ``good'' answer, while the other is designated as the ``bad'' answer. The detailed distribution of selected responses across models is shown in \figref{fig:mc_option_dist}.


\section{Experiments on \datamc}\label{sec:experiments_mc}

We evaluate the same models as in \secref{sec:experiment_gen}.
In our evaluation, we test models by presenting each question with its two corresponding options in alternating orders. This approach helps eliminate potential position bias in model responses while maintaining the fundamental binary choice structure. Models are prompted to select their preferred answer, and their performance is assessed based on their ability to consistently identify the better response.

\begin{table*}[t]
\centering
\small
\begin{tabular}{lS[table-format=3.2]S[table-format=3.2]S[table-format=3.2]S[table-format=3.2]S[table-format=3.2]}
\toprule
\textbf{Model} & \textbf{Good First} & \textbf{Bad First} & \textbf{Avg} & \textbf{Positional Bias} & \textbf{Format} \\
\midrule
      Llama-3.2-1B-Instruct &                   58.19 &                  39.35 &                48.77 &       18.84 &             59.68 \\
      Llama-3.2-3B-Instruct &                   65.43 &                  66.67 &                66.05 &       -1.24 &             53.99 \\
      Llama-3.1-8B-Instruct &                   76.97 &                  64.26 &                70.62 &       12.71 &             89.96 \\
     Llama-3.1-70B-Instruct &                   81.86 &                  75.04 &                78.45 &        6.82 &             98.67 \\
   Mistral-7B-Instruct-v0.1 &                   55.85 &                  46.96 &                51.41 &        8.89 &             99.70 \\
 Mixtral-8x7B-Instruct-v0.1 &                   69.22 &                  76.77 &                72.99 &       -7.55 &             96.23 \\
Mixtral-8x22B-Instruct-v0.1 &                   74.77 &                  71.71 &                73.24 &        3.07 &             97.93 \\
      Qwen2.5-0.5B-Instruct &                  100.00 &                   0.00 &                50.00 &      100.00 &             89.66 \\
        Qwen2.5-3B-Instruct &                   74.28 &                  58.98 &                66.63 &       15.30 &             87.22 \\
        Qwen2.5-7B-Instruct &                   68.59 &                  72.97 &                70.78 &       -4.38 &             53.99 \\
       Qwen2.5-32B-Instruct &                   77.00 &                  72.36 &                74.68 &        4.64 &             99.48 \\
       Qwen2.5-72B-Instruct &                   75.11 &                  74.70 &                74.91 &        0.41 &             99.78 \\
    claude-3-haiku-20240307 &                   73.41 &                  67.36 &                70.38 &        6.06 &            100.00 \\
   claude-3-sonnet-20240229 &                   67.21 &                  67.36 &                67.28 &       -0.15 &            100.00 \\
     gpt-4o-mini-2024-07-18 &                   72.23 &                  67.06 &                69.65 &        5.17 &            100.00 \\
          gpt-4o-2024-05-13 &                   81.22 &                  71.89 &                76.56 &        9.33 &             99.48 \\
\bottomrule
\end{tabular}
\caption{
\datamc evaluation results. \textbf{Good First} and \textbf{Bad First} are the accuracy in the percentage of selecting the correct answer where the correct answers are the first one and second one, respectively. \textbf{Avg} is the average of \textbf{Good First} and \textbf{Bad First}, with the random baseline of $50\%$. \textbf{Positional Bias} indicates the models' position bias to the first answer, and the closer it is to 0, the better. \textbf{Format} is the percentage of answers generated by the model in the correct format specified in the prompt.
}
\label{tab:average_score_se}
\end{table*}

\subsection{Main Results}\label{sec:experiments_res}
\Cref{fig:average_score_mc} shows the overall model performance on \datamc. In the multiple-choice evaluation setting, the general finding that larger models perform better still holds. We can observe models with large models in the Llama, Qwen, Mixtral family and \model{GPT-4o} achieved at least 40 scores in micro average across all categories of questions, which shows that they are significantly better than the random baseline. On the other hand, the ranking of the top-performing models has changed significantly. The best-performing model (\model{Claude-3-haiku}) in open generation evaluation ranks only in the middle tier of all models, while \model{Llama-3.1-70B} and \model{GPT-4o} now take the lead with micro average scores of 56.90 and 53.12, respectively. 

There are three small models \model{Mistral-7B}, \model{Qwen2.5-0.5B}, and \model{Llama-3.2-1B} struggle on the multiple-choice evaluation with almost zero performance difference compared to random baseline across all question categories. This observation suggests that these models cannot understand the concept and definition of trick questions and hence unable to accurately assess the answers to these questions, reaffirming that \datamc had the advantages of standardized evaluation and clear success criteria.

\begin{figure}[t]
    \centering
    \includegraphics[width=\linewidth]{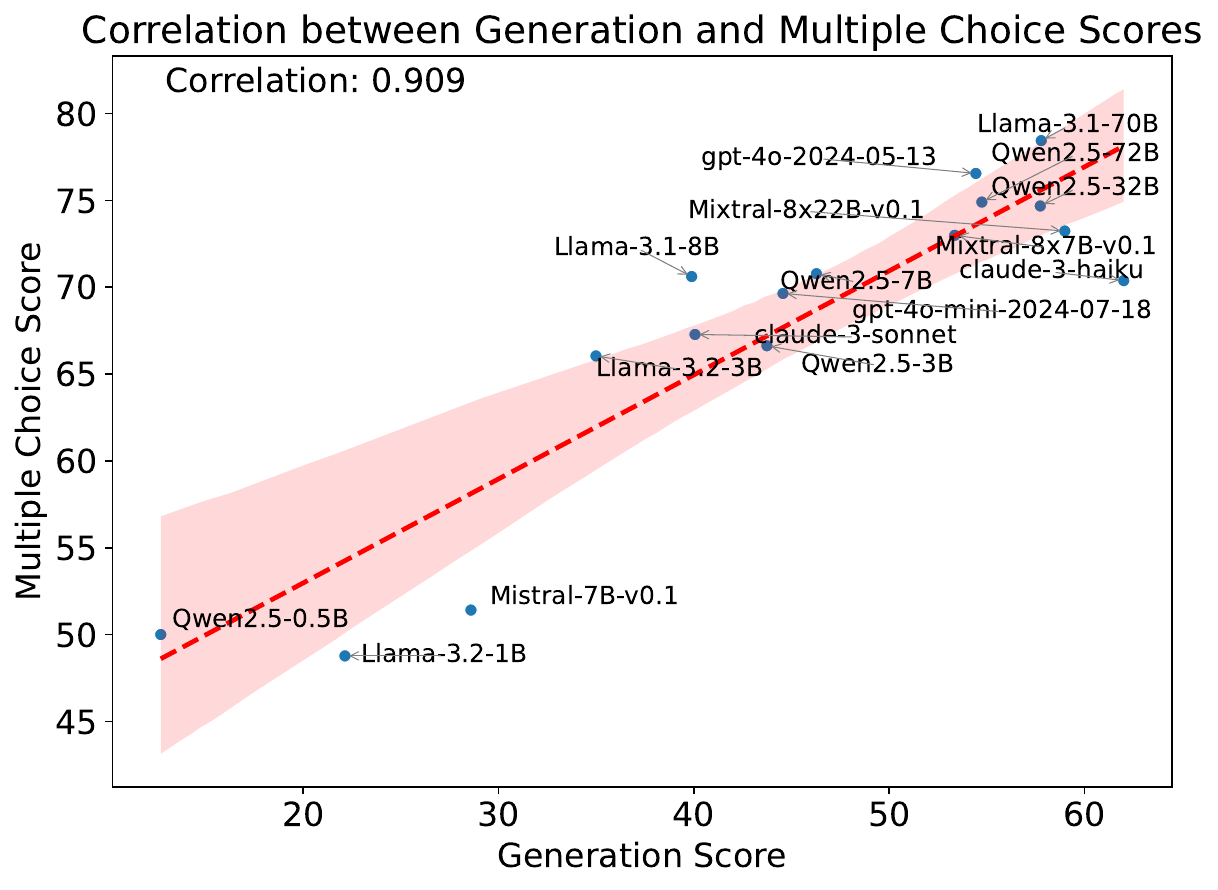} 
    \caption{Pairwise scatter plots with Pearson correlation coefficients of generation and multiple choice scores.}
    \label{fig:gen_mc_correlation}
\end{figure}

\begin{figure*}[t]
    \centering
    \includegraphics[width=\textwidth]{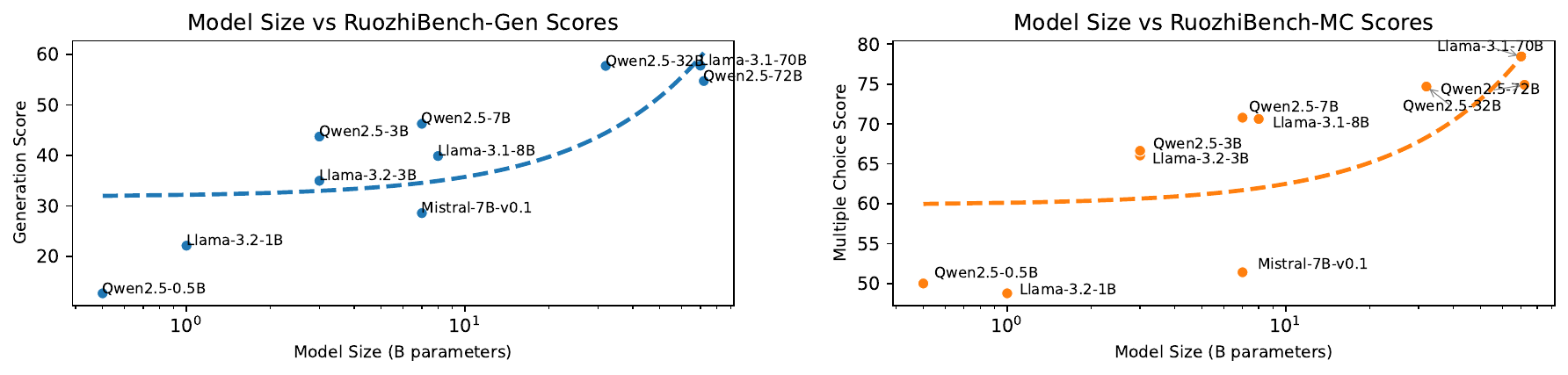} 
    \caption{
    Relationship between model size and performance on generation and multiple-choice tasks. The plots show the correlation between model size (in billions of parameters) and performance scores for both generation (top) and multiple-choice (bottom) tasks. Both plots use a logarithmic scale for model size. The dashed lines represent the regression fit, demonstrating a positive correlation between model size and performance for both task types.
    }
    \label{fig:size_score_correlations}
\end{figure*}

\subsection{Analysis}
\paragraph{Correlation with \datagen}
\Cref{fig:gen_mc_correlation} shows the correlation between generation and multiple choice scores for all models. We can observe a strong positive correlation between the generation and multiple choice scores for all models, with a Pearson correlation coefficient of 0.909. In general, most models have achieved slightly higher scores in the multiple choice than generation evaluation.

\paragraph{Model Size Analysis}
\Cref{fig:size_score_correlations} shows the relationship between model size and performance on generation and multiple-choice tasks.

\paragraph{Issues in MC}
Despite the advantages discussed above, we found two caveats of \datamc based on the detailed results in \Cref{tab:average_score_se}. (1) We found different degrees of performance gaps between when we provide the better response as the first option and the reverse, even for some of the best-performing models like \model{GPT-4o} and \model{Claude-3-haiku}. Most models perform slightly better when the better answer is provided as the first option. This \textbf{positional bias} suggests these models may be influenced by the ordering of options and indicates some uncertainty in their decision-making process.
(2) Not all models can strictly \textbf{follow the formatting} instructions we provided in the prompts of \datamc. Except for \model{Claude-3} models and \model{GPT-4o}, all other models produce different numbers of responses with formatting errors. Smaller models in \model{Llama-3.2} family and \model{Qwen2.5-7B} suffer more from this issue as their formatting success rates are less than 60\%.

\section{Related Work}\label{sec:related_work}

\paragraph{General Reasoning Evaluation of LLMs}
Evaluating the reasoning capabilities of LLMs has gained significant attention, with diverse benchmarks developed for different reasoning domains, such as commonsense reasoning \cite{commonsenseqa,hellaswag,arc,Bisk2020}, math \cite{gsm8k,math}, code \cite{humaneval, mbpp}, and logic \cite{liu2020logiqachallengedatasetmachine,logiqa2,liu2023evaluatinglogicalreasoningability}. 
Recent advances, with models like GPT-4 surpassing human performance on many of these benchmarks, have driven further exploration into more challenging testbeds. Models such as GPT-o1 \cite{openai2024openaio1card} and Deepseek-R1 \cite{deepseekai2025deepseekr1incentivizingreasoningcapability} have demonstrated improved performance on advanced benchmarks like AIME \cite{aime} and HLE \cite{phan2025humanitysexam}, which assess reasoning across domains such as mathematics, physics, and scientific knowledge. In contrast, \data presents seemingly simple questions---ones even a five-year-old could find fallacy---that expose fundamental gaps in LLMs’ commonsense reasoning abilities, highlighting the limitations of current models beyond factual knowledge and formulaic problem-solving.

\paragraph{Understanding Deceptive and Fallacious Texts}
While there is a substantial body of work on LLMs' reasoning capabilities, research specifically focused on evaluating how models handle deliberately deceptive or fallacious inputs remains limited. Recent work has begun exploring the use of Chinese \textit{Ruozhiba} forum data for improving LLMs' capabilities; for instance, \citet{lin2024baichuanalignmenttechnicalreport} and \citet{bai2024coigcqia} incorporated \textit{Ruozhiba} data in their training data to enhance logic reasoning in Chinese. 

There are several works exploring LLMs' understanding of logical fallacies \cite{lei-huang-2024-boosting,payandeh2023susceptiblellmslogicalfallacies,li2024reasonfallacyenhancinglarge}. While most relevant work is \citet{li2024when}, which created a benchmark using data from \textit{Ruozhiba}. However, our work differs in that: (1) we provide the first English benchmark, while theirs is Chinese-only; (2) their evaluation relies on artificially-constructed input formats, whereas our evaluation setting is more natural, directly using questions as prompts; and (3) we include detailed annotations of fallacy types, enabling more systematic analysis of model capabilities.
Through these innovations, we aim to enable more rigorous assessment of how LLMs handle the types of deliberately tricky or misleading inputs they may encounter in real-world applications.

\section{Conclusion}\label{sec:conclusion}
This paper presents \data, a comprehensive benchmark designed to evaluate the logical reasoning capabilities of LLMs through both generative and multiple-choice formats. Our analysis across diverse models reveals that while state-of-the-art models like Claude demonstrate strong performance on logical reasoning tasks, significant challenges remain, particularly in handling edge cases and complex logical structures. The dual format of our benchmark provides complementary insights into models' reasoning abilities, suggesting several promising directions for future research, including the enhancement of model training and the development of more targeted approaches to improving logical reasoning capabilities.

\section*{Limitations}
Despite our efforts to create a comprehensive benchmark for logical reasoning, \data has several limitations. First, while our multiple-choice format offers standardized evaluation, it may not fully capture the nuanced reasoning processes that models employ in real-world scenarios. Second, our evaluation method relies heavily on model-generated responses for constructing the trapped options, which might not encompass all possible fallacies or reasoning errors that LLMs could make. Additionally, although the dataset is bilingual, our experiments focus primarily on English. Finally, the binary choice format in \data-MC, while effective for evaluation, may inadvertently simplify complex reasoning problems that in practice require consideration of multiple valid perspectives or solutions.


\bibliography{custom}

\clearpage
\newpage
\onecolumn

\appendix

\section{Prompts Used in This Study}
\begin{figure}[h]
	\small
	\begin{tcolorbox}[colframe=white, left=2.5mm, right=1.5mm]
Please read the following question and point out the irrationality in the question based on correct knowledge and common sense. The answer should be concise. (Note: Do not answer this question, do not use words like "the irrationality of this question is", your output only needs to include the irrationality of the question, try to use one sentence to complete the answer, and the answer should not exceed 100 words.) \\

Example:\\
Question: If the sun rises at night, what impact will it have on the temperature of the Earth?\\
Irrationality Analysis: The sun does not rise at night because day and night are caused by the rotation of the Earth, and the phenomenon of the sun rising and falling is the result of the Earth's rotation. Assuming that the sun rises at night is contrary to basic astronomical knowledge.\\

Inputs:\\
Question: \blue{\{question\}}
	\end{tcolorbox}
	\caption{Irrationality analysis generation prompt.}
	\label{fig:trick_gen}
\end{figure}

\begin{figure*}[h]
	\small
	\begin{tcolorbox}[colframe=white, left=2.5mm, right=1.5mm]
Based on the following \textit{tricky question} and the \textit{irrationality analysis} of this question, analyze and label them with three closest question categories. You will see all question categories in the question classification criteria, and you need to output the number sequence of question categories according to priority.\\

Question Classification Criteria:\\
1. Logical error: When the question is raised, there may be logical contradictions or reasoning errors, which may include violations of logical rules, such as informal or formal logical errors, making the question logically untenable.\\
2. Common sense misunderstandings: The question reflects a misunderstanding of some basic common sense or universally accepted facts, usually involving incorrect understanding of common knowledge or information in daily life.\\
3. Erroneous assumption: The question is based on one or more erroneous assumptions. This means that the foundation or starting point of the question is incorrect, leading to inaccuracies in the entire question or its answer.\\
4. Scientific misconceptions: The content of the question involves misunderstandings of scientific principles or knowledge, including incorrect interpretations or applications of scientific theories, experimental results, or scientific methods.\\
5. Absurd imagination: The situation or setting in the question is clearly contrary to reality or common sense, has absurdity, and usually contains impossible or illogical elements, making the entire question appear unrealistic.\\
6. Others: If you believe that the provided question categories do not match the current question, please choose this option.\\

Definitions:\\
- Tricky Question: A question that is illogical or lacks a clear answer, often based on unrealistic assumptions or contradictions to common sense.\\
- Irrationality Analysis: A detailed explanation highlighting the logical flaws or unrealistic assumptions present in the tricky question.\\

Output format:\\
The output format must be *Number Space Number Space Number*, and the output string length must be 5.\\
Example: 2 6 3\\

Inputs:\\
Question: \blue{\{question\}}\\
Irrationality Analysis: \blue{\{irrationality\}}
	\end{tcolorbox}
	\caption{Question type classification prompt.}
	\label{fig:qc}
\end{figure*}

\begin{figure*}[t]
	\small
	\begin{tcolorbox}[colframe=white, left=2.5mm, right=1.5mm]
You are an evaluation system designed to assess responses to *deceptive* or *logically flawed* questions. \\

\#\# Task Definition:\\
For each input, you will be provided with:\\
- A **deceptive question** that contains misleading premises, logical fallacies, or contradictions.\\
- An **irrationality analysis**, explaining why the question is deceptive and identifying its logical flaws.\\
- An **answer** provided by a model or a human.\\

Your task is to **evaluate the answer's ability to correctly identify and address the deception** rather than simply providing a direct response.\\

\#\# Rating Criteria (0-4 Scale):\\
- **0**: The answer completely fails to recognize the deception and directly attempts to answer the question as if it were valid.\\
- **1**: The answer acknowledges a flaw but does not engage with it meaningfully.\\
- **2**: The answer recognizes and partially explains the deception but has logical gaps.\\
- **3**: The answer correctly identifies and mostly explains the fallacy but lacks depth or clarity.\\
- **4**: The answer fully explains the deceptive nature, providing a comprehensive and logically sound explanation.\\


<question>\verb|\n|\blue{\{question\}}\verb|\n|</question>\\

<irrationality\_analysis>\verb|\n|\blue{\{irrationality\_analysis\}}\verb|\n|</irrationality\_analysis>\\

<answer>\verb|\n|\blue{\{answer\}}\verb|\n|</answer>\\

\#\# Output Format:\\
You must return a JSON object with keys: "rating" (int) and "explanation" (str), for example:\\
\{\{\verb|\n|"rating": [2],\\
"explanation": "[Your justification for the rating]"\verb|\n|\}\}\\
Ensure that your response is **valid JSON** with the exact keys shown above without any additional text or formatting.
	\end{tcolorbox}
	\caption{\datagen evaluation prompt.}
	\label{fig:eval_prompt}

    \begin{tcolorbox}[colframe=white, left=2.5mm, right=1.5mm]
You are an expert evaluator tasked with assessing the quality of an answer to a given question. \\

\#\# Question:\verb|\n|\blue{\{question\}}\\

\#\# Answer:\verb|\n|\blue{\{answer\}}\\

\#\# Evaluation Criteria:\\
Rate the answer based on the following criteria:\\
- 4: Excellent – The answer is complete, accurate, and well-explained.\\
- 3: Good – The answer is mostly correct with minor inaccuracies or missing details.\\
- 2: Fair – The answer has some correct elements but contains notable errors or omissions.\\
- 1: Poor – The answer is mostly incorrect or incomplete.\\
- 0: Very Poor – The answer is irrelevant or completely incorrect.\\

\#\# Output Format:\\
You must return a JSON object with keys: "rating" (int) and "explanation" (str), for example:\\
\{\verb|\n|"rating": [0, 1, 2, 3, or 4],\\
"explanation": "[Your justification for the rating]"\verb|\n|\}\\
Ensure that your response is **valid JSON** with the exact keys shown above without any additional text or formatting.\\
	\end{tcolorbox}
	\caption{\datamc evaluation prompt.}
	\label{fig:normal_eval_prompt}
\end{figure*}

\clearpage
\section{Option Distribution of \datamc}
\begin{figure*}[h]
    \centering
    \includegraphics[width=\linewidth]{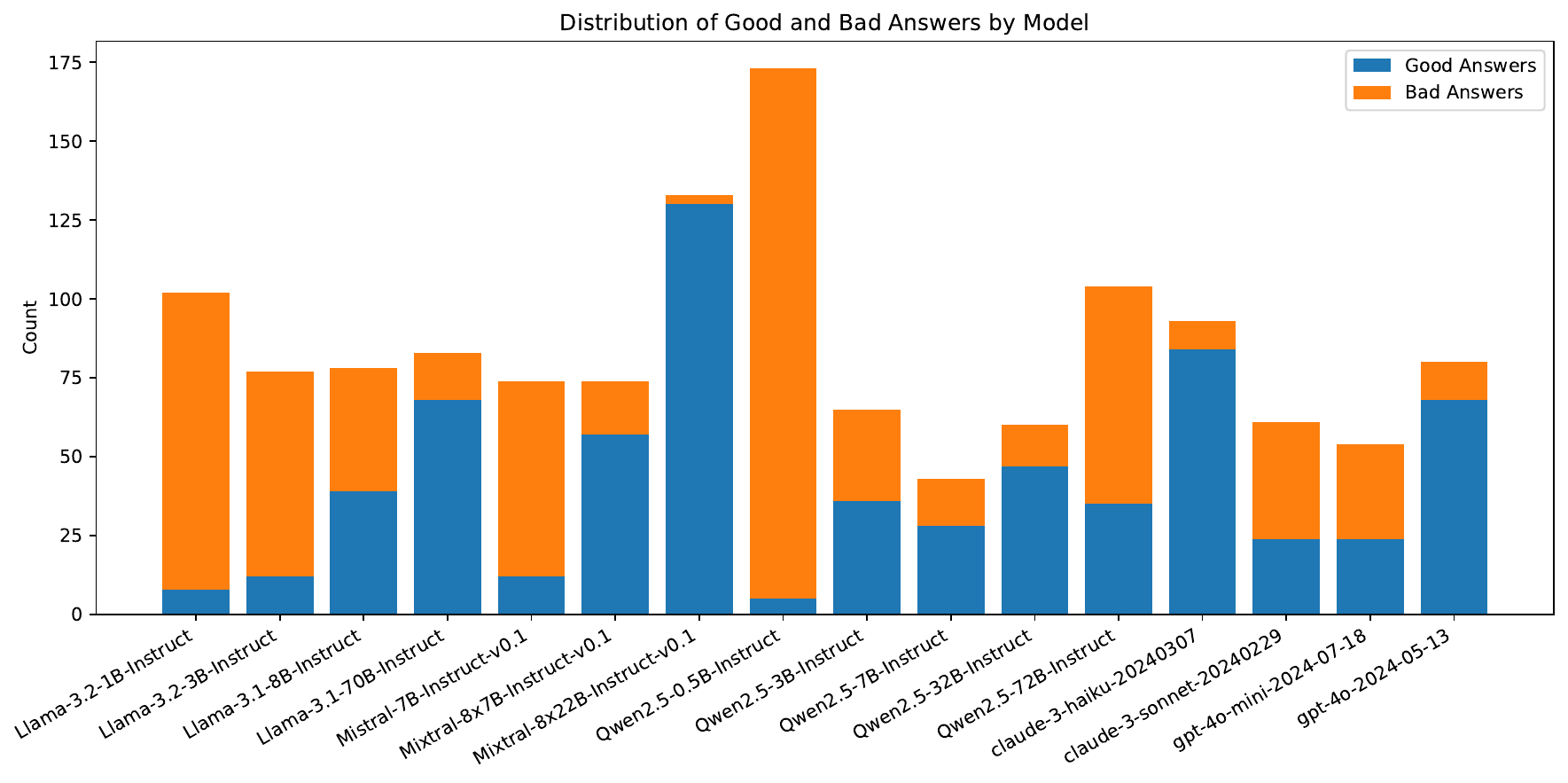} 
    \caption{Distribution of Good and Bad Answers by Model. The figure shows the total number of responses across various models, divided into good and bad answers. Most models exhibit a relatively balanced distribution, while models like Claude 3 Sonnet, Mixtral 8x22B, and GPT-4o produce a higher proportion of good answers. In contrast, models like Qwen 2.5 0.5B have a substantial number of responses but with a higher proportion of bad answers.}
    \label{fig:mc_option_dist}
    
    \includegraphics[width=.8\linewidth]{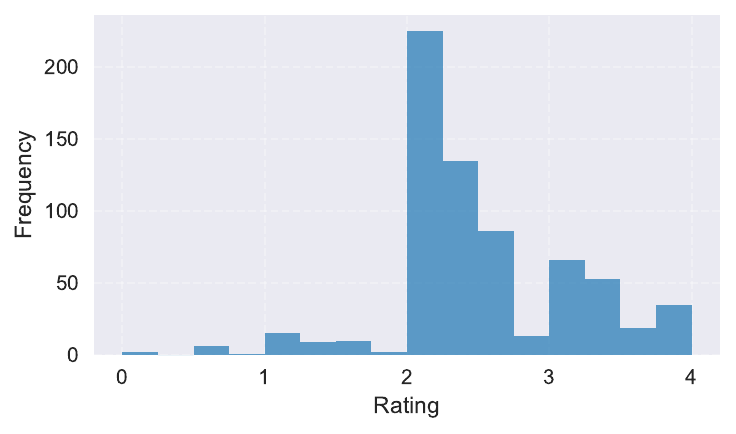} 
    \caption{``Good'' and ``Bad'' answer scores distribution. The majority of the data falls into categories with score differences greater than 2, indicating a clear gap between the options.}
    \label{fig:mc_gap_dist}
\end{figure*}

\figref{fig:mc_option_dist} shows the option data source statistics of \datamc, and \figref{fig:mc_gap_dist} shows the gap distribution between ``Good'' and ``Bad'' options. 

\section{Recruitment and Payment}\label{sec:annotator}
We hired 2 annotators with bachelor's degrees or higher from China with an hourly rate of 50 Chinese Yuan. The annotators are native Chinese speakers and have studied English for more than 10 years. This rate is higher than the average hourly wage in China.

\newpage
\clearpage

\section{Full Evaluation Results on \datagen}\label{sec:all_eval}

\begin{figure}[ht]
    \centering
    \includegraphics[width=0.7\linewidth]{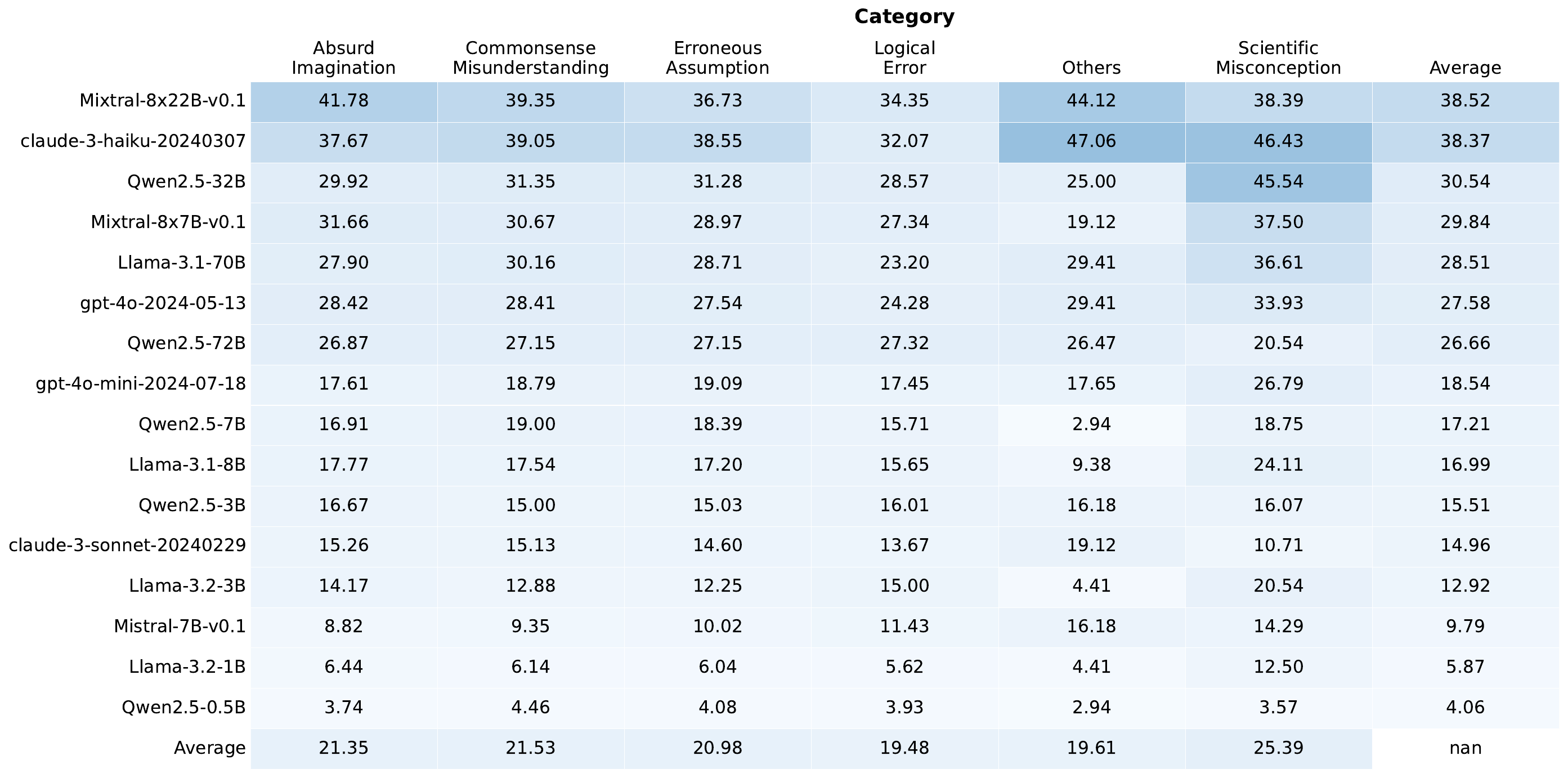}
    \caption{Overall score on \datagen using \model{Claude-3-5-sonnet} as an evaluator.}
    \includegraphics[width=0.7\linewidth]{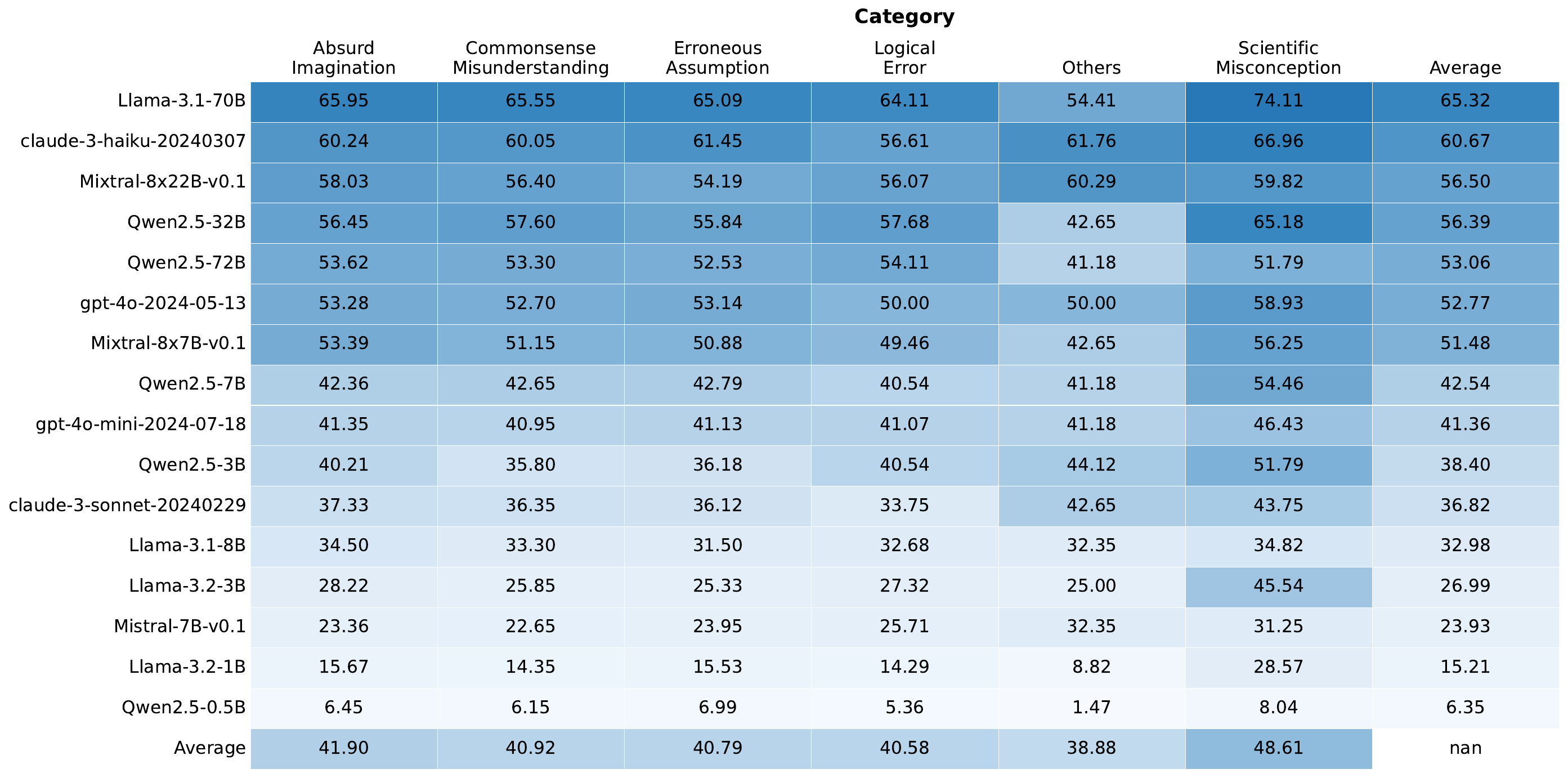}
    \caption{Overall score on \datagen using \model{GPT-4o-2024-08-06} as an evaluator.}
    \includegraphics[width=0.7\linewidth]{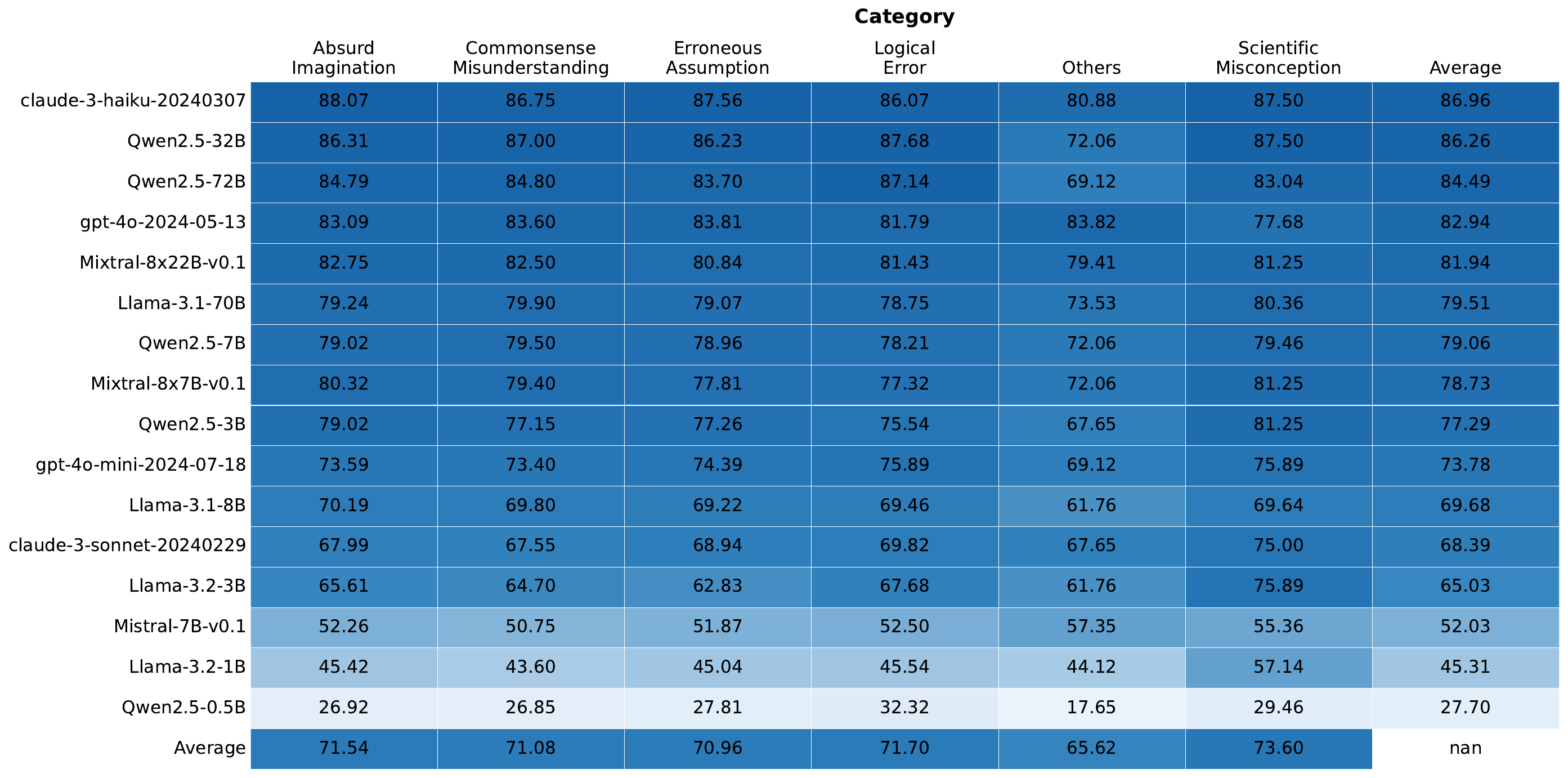}
    \caption{Overall score on \datagen using \model{Llama-3.3-70B-Instruction} as an evaluator.}
    \label{fig:all_eval}
\end{figure}

\clearpage
\section{Rating Distribution of Evaluators on \datagen}

\begin{figure}[ht]
    \centering
    \includegraphics[width=0.7\linewidth]{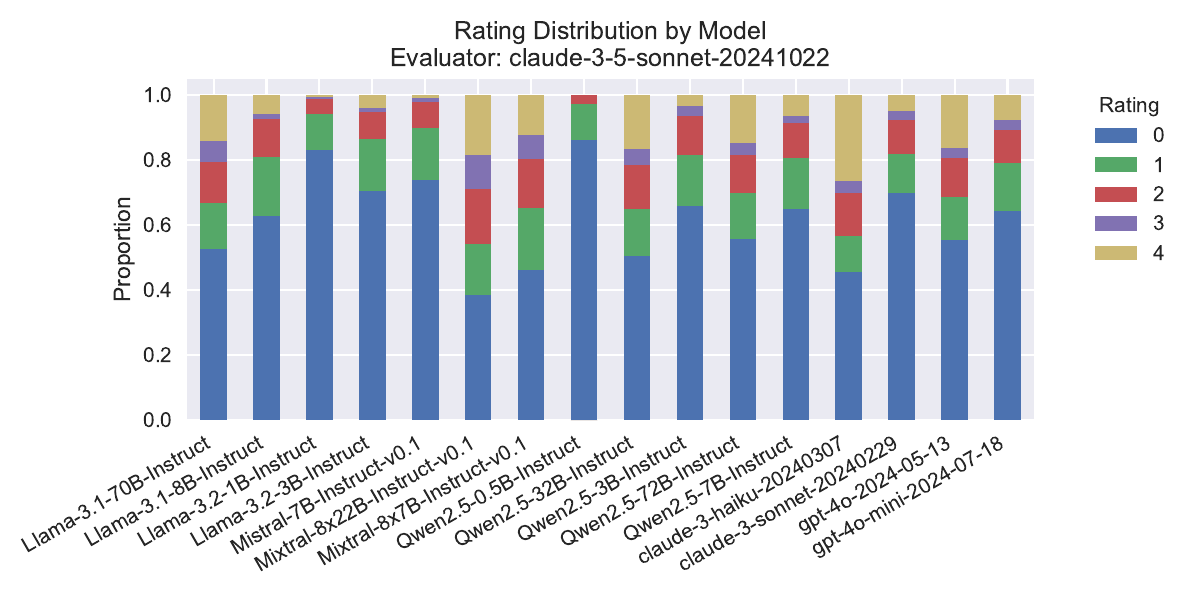}
    \caption{Rating distribution on \datagen using \model{Claude-3-5-sonnet} as an evaluator.}
    \includegraphics[width=0.7\linewidth]{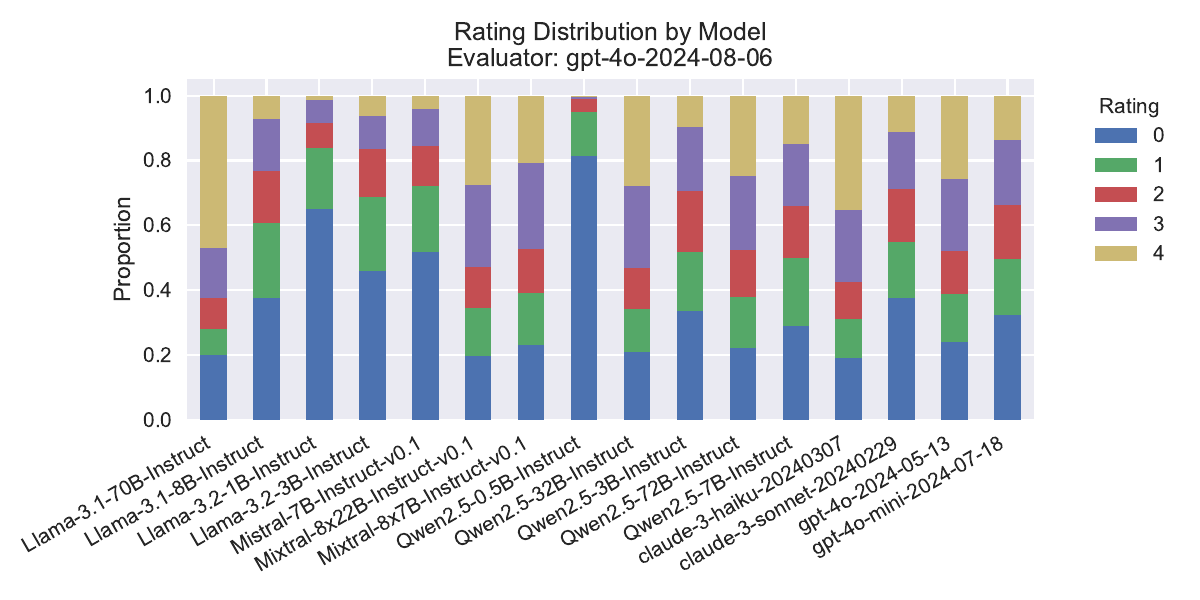}
    \caption{Rating distribution on \datagen using \model{GPT-4o-2024-08-06} as an evaluator.}
    \includegraphics[width=0.7\linewidth]{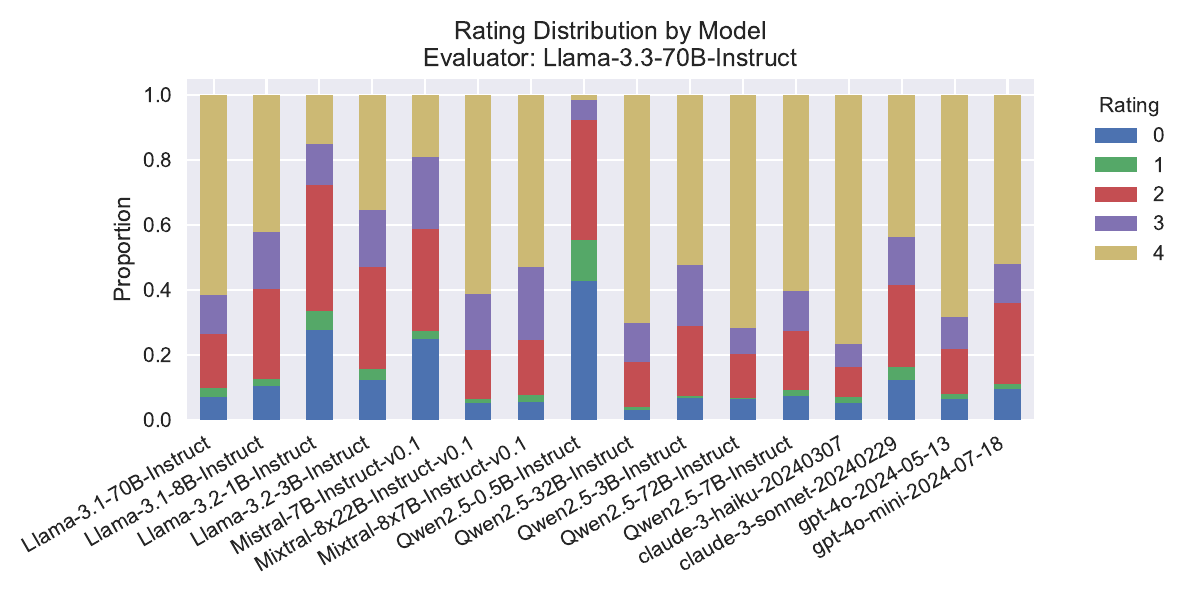}
    \caption{Rating distribution on \datagen using \model{Llama-3.3-70B-Instruction} as an evaluator.}
    \label{fig:all_rate}
\end{figure}

\end{CJK*}
\end{document}